\documentclass[manuscript,screen]{acmart}
\AtBeginDocument{%
  }

\setcopyright{acmlicensed}
\copyrightyear{2027}
\acmYear{2027}
\acmDOI{XXXXXXX.XXXXXXX}
\acmConference[CSCW '27]{TBD}{}{, }
\acmISBN{978-1-4503-XXXX-X/2018/06}
\usepackage{multirow}
\usepackage[table]{xcolor}

\begin{document}

\title[When Youth Enter The Chat]{When Youth Enter The Chat: An Epistemic Shift in the Validation of LLM-Based Measures of Student Talk}

\author{Liliana Santos-Deonizio}
\affiliation{%
  \institution{Stanford University}
  \city{Stanford, CA}
  \country{United States}}
\email{lilianas@stanford.edu}

\author{James Malamut}
\affiliation{%
  \institution{Stanford University}
  \city{Stanford, CA}
  \country{United States}}
\email{jmalamut@stanford.edu}

\author{Ram\'{o}n Antonio Mart\'{i}nez}
\affiliation{%
  \institution{Stanford University}
  \city{Stanford, CA}
  \country{United States}}
\email{ramon.martinez@stanford.edu}

\author{Dorottya Demszky}
\affiliation{%
  \institution{Stanford University}
  \city{Stanford, CA}
  \country{United States}}
\email{ddemszky@stanford.edu}

\renewcommand{\shortauthors}{}

\begin{abstract}
Large Language Models (LLMs) are being used increasingly to measure aspects of student discourse (e.g. talk moves, collaboration, equity of voice) at scale. Typically, LLM-based measures of student talk use transcriptions of classroom conversations that only include verbal contributions, which de-contextualize student language. Common practices for validating these measures include comparing outputs against expert annotations by adults, using held out evaluation sets and F1 scores. We argue that these approaches are insufficient to ensure that such measures are meaningful and equitable for teaching and learning, particularly for racially and linguistically marginalized youth. In order to center the youth whose talk is being analyzed, re-contextualizing these classroom conversations and engaging youth in the research process is necessary. Sharing epistemic authority with youth, ultimately, centers their point of view and adds crucial nuance to the analysis of their talk that adult experts, researchers, and LLMs cannot provide. In a case study of multilingual youth in one 8th-grade math classroom, we address the epistemic exclusion of youth by employing multiple ethnographically-oriented methods to re-contextualize student conversations and center youth as epistemic authorities in conversation with researchers and LLMs. We conducted participant observations, interviews, focus groups, and member checks with four focal students. Findings reveal that there were misalignments between students' interpretations of their own math talk experiences and the LLM-based measures of their talk. Students contested both the LLM classifications and the coding scheme used to measure their talk, highlighting the need for youth to be involved in the epistemic process of producing knowledge about their experiences.

\end{abstract}




\maketitle

\begin{table*}[h!]
\centering
\caption{Excerpt 1 from Classroom Conversation between Ximena and Destiny}
\label{tab:excerpt1}
\begin{tabular}{p{0.12\textwidth} p{0.08\textwidth} p{0.72\textwidth}}
\toprule
\textbf{Speaker} & \textbf{Line} & \textbf{Dialogue} \\
\midrule
Ximena  & 1 & ``But I subtract th-, that, that to that.'' \\
Destiny & 2 & ``No, you minus on these sides, right?'' \\
Ximena  & 3 & ``Yeah.'' \\
Destiny & 4 & ``That's how we learned yesterday, and then do the same thing on these numbers once you figure out this answer.'' \\
Ximena  & 5 & ``Oh, oh it's ninety one. That's why.'' \\
        & 6 & ``This is ninety one.'' \\
        & 7 & ``Ninety one divided by nine.'' \\
        & 8 & ``It's why we don't do math.'' \\
        & 9 & ``Math pisses me off.'' \\
\bottomrule
\end{tabular}
\end{table*}

\section{Introduction}
"It's why we don't do math. Math pisses me off." Ximena, an 8th-grade Mexican American girl, said to her friend and classmate, Destiny, an 8th-grade Dominican and Haitian American girl. They were working on a problem and Ximena had asked Destiny for help. She expressed her frustration with the process of solving the problem. All the while, two small microphones captured their words and an iPad filmed while they worked. Later, this conversation would be transcribed (automatic transcription reviewed and edited by human transcribers) and an LLM would be used to annotate it for mathematic talk moves. These two sentences from Ximena would be coded as "Off-task," meaning they did not relate to the math task at hand or the process of solving the problem. This definition of Off-task had come from educational research about math and collaboration, and it distinguished other academic talk moves from Off-task talk. A few weeks later, the first author sat with Ximena to watch a clip of the exchange.  When asked whether she agreed that those lines were off-task, Ximena pushed back: "Technically I wasn’t really off task, I was talking about math but not in a good way… So, Chat PT [\emph{sic}] needs to get his life together.”

\begin{figure}[h!]
  \centering
  \includegraphics[width=.5\linewidth]{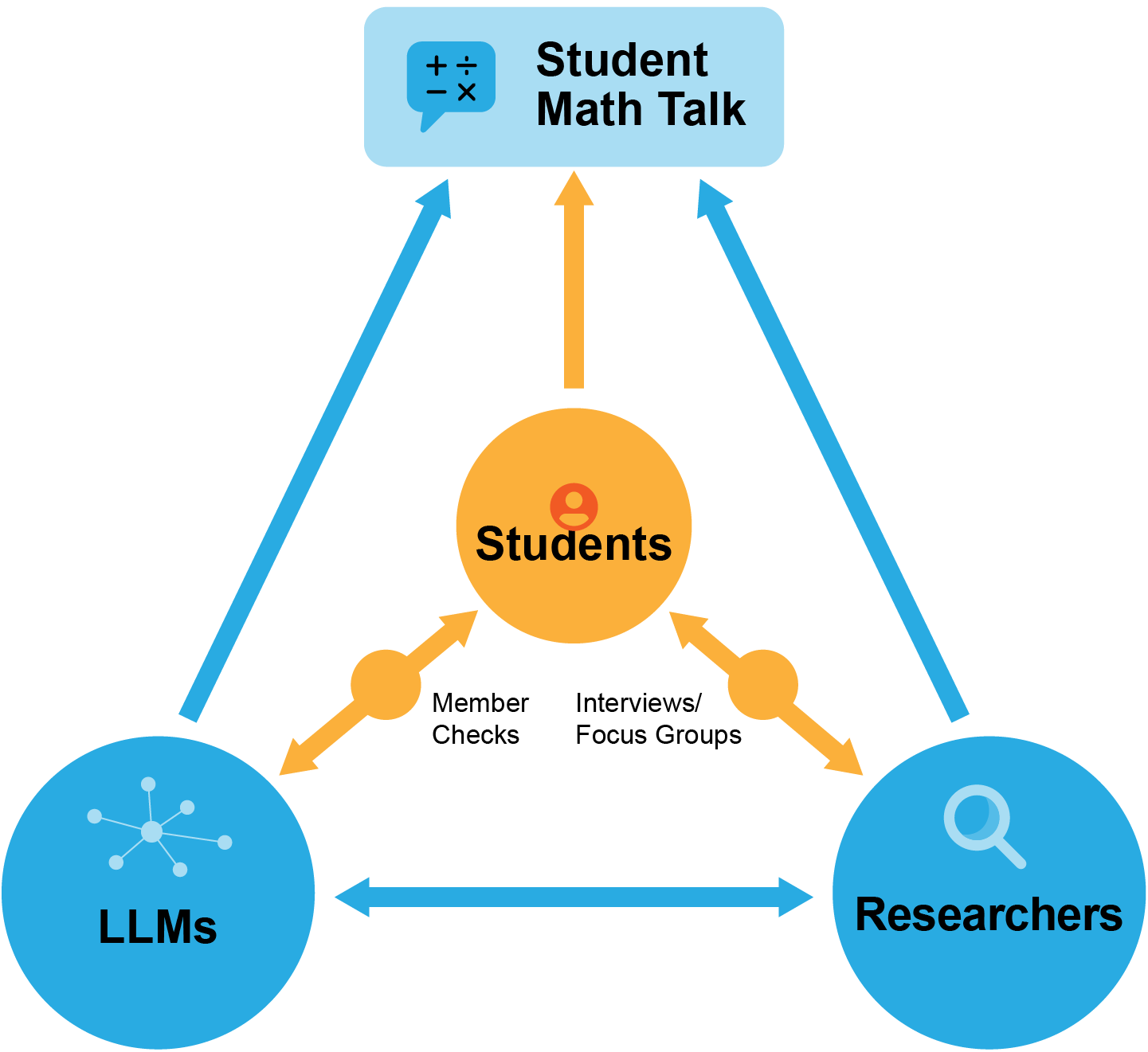}
  \caption{When Youth Enter the Chat: Before (blue) and After (yellow) Involving Youth in the Knowledge Production Process \label{fig:diagram}}
  \Description{student map}
\end{figure}

In this paper we explore the following question: \textbf{what can be learned by involving students in the epistemological process of validating LLM-based measures of student talk?} Validation itself is a practice of knowledge production, and it raises the question of who holds \emph{epistemic authority} over how student language is interpreted. Students are experts with respect to their own experiences and bringing them into the validation process can deepen the understanding of their talk in ways that teachers, researchers, and LLMs cannot achieve alone. We argue for a two-part \emph{epistemic shift}. First, language should not be treated solely as a text-based representation, but as a situated, social, cultural, and embodied experience --- which requires \emph{re-contextualizing} the transcripts that LLM-based measures rely on. Second, youth should be brought into the validation process as epistemic authorities on their own talk, not just as the subject of expert annotation. Figure~\ref{fig:diagram} maps this shift from excluding to including students in the validation process.

We illustrate this shift through a case study of multilingual youth\footnote{We use the term multilingual youth expansively to refer to youth that speak more than one language or language variety, i.e. African American English or Spanish-English varieties \cite{martinez_black_2022}.} in one 8th-grade math classroom, drawing on participant observations, interviews, focus groups, and member checks with four focal students. Re-contextualizing the transcripts surfaces dimensions of participation — relationships, classroom routines, physical space, gesture, prosody, language ideologies — that text-only representations obscure and that shape potential interpretations of student talk. We also show how sharing epistemic authority with the students reveals misalignments between how the LLM classified their talk and how they themselves understood it. These misalignments go beyond errors that can be fixed with prompt-engineering and instead require modifications of the coding scheme that delineates what counts as math talk. Together, the two shifts point toward a model of validation in which youth are not only the source of the data, but also participants in deciding what the data means.

The remainder of the paper illustrates these two shifts. We start in Section~\ref{sec:related_work} by reviewing related work on computational measures of classroom discourse, ethnographically oriented methods and participatory approaches to AI evaluation. Next, in Section~\ref{sec:llm_measures}, we provide a brief overview of the LLM-based measures for context. In Section~\ref{sec:ethnographic_methods}, we describe the ethnographic methods we used for re-contextualization and member checking. Section~\ref{sec:results} presents what each surfaces, including the context the transcripts miss and classifications the students themselves contest.

\section{Related Work}
\label{sec:related_work}

\subsection{Computational Measures of Classroom Discourse}
Recent work in Natural Language Processing (NLP) and learning analytics has used computational methods to analyze student and teacher discourse, including talk moves, collaboration, and patterns of participation in classroom and collaborative problem-solving settings \cite{demszky-hill-2023-ncte, pugh_say_2021, pugh_speech-based_2022, reitman_multi-theoretic_2023, anderson_exploring_2025, park_understanding_2025, meyer_using_2024, ahtisham_optimizing_2026}. Prior studies have used methods such as Linguistic Inquiry and Word Count (LIWC) \citep{park_understanding_2025, pugh_speech-based_2022}, BERT \citep{pugh_speech-based_2022, pugh_say_2021}, and other classifiers to identify discourse features and relate them to learning outcomes or collaborative behaviors. These studies have advanced the ability to analyze classroom talk at scale and have typically validated models using expert-annotated transcripts, held-out test sets, and performance metrics such as F1 scores, AUROC, or agreement with human experts. While these approaches are useful for assessing how well models reproduce expert labels, they generally treat transcripts as sufficient representations of classroom interaction and rarely incorporate validation from the perspectives of the students whose discourse is being measured. As a result, current validation practices say little about whether model outputs align with how students themselves understand their participation or with the broader contexts in which that talk occurs.

\subsection{Ethnographically-Oriented Methods and Context in Multilingual Math Classrooms.}
What ethnographic research, in particular, has long shown is that classroom discourse is shaped by social relationships, institutional norms, identity, and embodied interaction. Studies focused on multilingual youth translanguaging practices \citep{wei_not_2022, garcia_translanguaging_2014, poza_language_2018}, and language ideologies \citep{martinez_reading_2013, flores_producing_2024, flores_undoing_2015} demonstrate that meaning cannot be understood from words alone, but must be interpreted in relation to context, audience, language ideologies and the racialized experiences of interlocutors. The context of multilingual mathematics classrooms makes these limitations especially important. In the United States, the language practices of multilingual youth have historically been framed through a deficit lens \cite{gutierrez_at_2006}, positioning their linguistic repertoires as requiring remediation rather than serving as resources for learning \cite{gonzalez_funds_2009, rosa_looking_2018}. Research in mathematics education has shown that talk and collaboration are central to learning, \cite{erath_designing_2021}, while scholarship on identity and language in math classrooms has demonstrated that participation is shaped by broader histories of race, gender, language, and belonging \cite{ortiz_black_2021, ortiz_lessons_2024, morales_jr_underlife_2025, gholson_restoring_2019, esmonde_power_2013}. In classrooms where standardized English is often treated as the norm \cite{charity_hudley_understanding_nodate, hill_language_1998}, multilingual students may draw on broad linguistic repertoires across settings \cite{garcia_translanguaging_2014}, and their participation may not be fully legible through transcript-based measures alone. This makes multilingual math classrooms a particularly revealing site for examining what LLM-based discourse measures capture, what they miss, and what is at stake if youth are not involved in the validation process.

\subsection{Participatory Approaches to AI Evaluation}
We engage youth through participatory approaches similar to work done in AI and HCI evaluation. Related work in HCI and responsible AI has argued that evaluating AI systems requires attention to the perspectives of the communities most affected by them \cite{harvey_dont_2025, solyst_i_2023, solyst_investigating_2025, tanksley_ethics_2025}. Participatory approaches, including audits \cite{solyst_investigating_2025}, co-design and co-creation \cite{morales-navarro_building_2025}, teacher and youth centered evaluation \cite{harvey_dont_2025, tanksley_ethics_2025}, have been used to surface harms, biases, and mismatches that are not visible through standardized metrics alone. Additionally, youth have been engaged through Youth Particpatory Action Research to reflect on technology and its impact on students \cite{gak_reflecting_2024}, and college students have been engaged in participatory design processes \cite{navarro_re-imagining_2022}. Our work builds on the evaluation strand of this literature: we ask what happens when youth are brought into the validation of LLM-based measures that researchers have already built — a stage in the AI lifecycle where their perspectives remain especially rare. 

\subsection{Epistemic Authority and Exclusion}
Finally, we draw on theoretical frameworks of epistemic agency, authority and exclusion to analyze what is missed when students are not involved in the validation process. \emph{Epistemic agency} is the ability for youth to participate in the disciplinary practices and intellectual activities of knowledge production \cite{mathayas_re-indexing_2026}. \emph{Epistemic authority} relates to who has the power to engage in the intellectual process of knowledge building, which traditionally has been held by teachers in classrooms, and scholars and their tools in research. \emph{Epistemic exclusion} relates to who is excluded from the knowledge-building process \cite{baze_call_2025}.  Scholars in math and science education have used these constructs to examine power and participation in classroom collaboration \cite{langer-osuna_so_2020, krist_teacher_2023}, showing that epistemic agency is dynamic and interactionally constituted, and that authority does not redistribute automatically in collaborative or student-centered settings. Additionally, scholars have applied this framework to show how technology can also mediate epistemic injustice \cite{ajmani_power_2025}.

We extend this body of work in two ways. First, while these frameworks have largely been applied to classroom interactions, we apply them to the research process itself. We ask how youth are epistemically excluded not only when learning, but also when their learning is being measured, interpreted, and represented. Second, building on work that positions youth as philosophers of technology \cite{vakil_youth_2022}, we engage youth as experts of their own experiences who can interrogate technological artifacts, in this case the LLM classifications of their talk.

\section{LLM-Based Measures of Student Talk}
\label{sec:llm_measures}

 The context of our study is a two-year project, one goal of which has been to develop scalable measures of multilingual students' math language practices. Data from the first year of the project (2024--2025 school year) was used to develop and validate preliminary LLM-based measures via prompt-engineering. We treat these LLM measures as central artifacts in the ethnographic work presented in this paper, both for re-contextualization and member-checking. The ethnographic work was conducted during the 2025--2026 school year. Here we provide a short overview of these LLM-based measures and their initial validation.
 
  In the first project year, the research team partnered with two large urban school districts in the West Coast, both serving a high number of multilingual students, to collect transcripts from middle school math classrooms. During recordings, students worked on math tasks with peers, with at least one student per group reporting Spanish as a home language. Recordings were transcribed by automated speech recognition software with human cleanup and review prior to annotation.

 \paragraph{Codebook.} We used a multi-theoretic coding scheme that covered topic and talk move codes from collaborative problem solving (CPS) and mathematics learning frameworks \cite{planas_quality_2021, pugh_say_2021, webb_engaging_2014, reitman_multi-theoretic_2023, drageset_different_2015}. LLMs were used to code for collaborative and academic talk moves, including the topic codes---Offtask, Recording, Competent, Understanding, Tool---and the talk move codes---Question, Apology, Claim, Reasoning, Next Step, Monitor, Disagree, Agree, Compare, Add-On, and Revoice. The codebook definitions were developed with human experts who also helped create the validation set. In this paper, we focus on the seven codes with the highest frequency in our analytic sample: Off-task, Understanding, Recording, Question, Claim, Next-step, Disagree. This analytic codebook is shown in Table~\ref{table:code_definitions}.

 \begin{table*}[h]
\centering
\caption{Code Definitions and Examples}
\label{table:code_definitions}
\begin{tabular}{p{0.10\textwidth} p{0.12\textwidth} p{0.40\textwidth} p{0.33\textwidth}}
\toprule
\textbf{Code Type} & \textbf{Code} & \textbf{Definition} & \textbf{Example} \\
\midrule

\multirow{2}{*}{Topic}
& Off-Task
& Talk not pertaining to the assigned task and how to engage with it.
& ``I'm just gonna go back there and play Fortnite. I somehow know how to play Fortnite.'' \\

& Understanding
& Discussing group or individual's understanding of a mathematical idea, question, or problem. Disscussing whether or not or to the extent to which the group or individual feels like they understand or do not understand.
& ``Okay, ya lo entiendo.'' (Okay, I understand it now) \\

& Recording
& Talk pertaining to the recording equipment or the university researchers using it to observe them. 
&  ``Estoy diciendo cosas y tengo el micrófono.'' (I am saying things and I have the microphone.)\\

\midrule

\multirow{5}{*}{Talk Move}
& Question
& Asks a question
& ``Why?''\\

& Claim
& A mathematical statement of fact or conjecture that can be proven false or not.
& ``Veinte por cuarenta es ochenta.'' (Twenty times forty is eighty.) \\

& Next-Step
& Suggests the next step for the group or an individual to take.
& ``Try and see if eight can go into twenty-four on this.'' \\

& Disagree
& Expresses disagreement with another students’ mathematical reasoning expressed in a prior turn. 
& ``No, you can't put the five because you never know how many friends there is.'' \\

\bottomrule
\end{tabular}
\end{table*}

 \paragraph{Validation set.} To develop a validation set for the LLM models, five transcripts totaling 41,196 utterances were annotated by human experts---current and former teachers and educators, including members of the research team. Specifically, the annotators were one Latina woman (former teacher and first author of this paper), one Asian woman (current curriculum specialist), three white women (a current teacher, an instructional coach, and a teacher educator), and one white man (former teacher and second author of this paper). Utterances were split at the sentence level that served as the unit of analysis for the LLM-based talk measures. Each utterance was additionally tagged for whether the utterance was in English, Spanish or both. We evaluated several LLMs against this validation set (Claude 3.7, GPT 5.0, GPT 5.1) and selected GPT-5.1, which performed best based on F1 scores. 

 \paragraph{LLM prediction.} We used GPT-5.1 to annotate the transcripts chosen for the student member checks. The model was prompted with the code definitions and three positive examples and three non-examples per code (chosen by human experts), and instructed to assign each utterance a binary (0/1) label for each code. The analytic sample consisted of two transcripts from the focal classroom (2025--2026).

 \paragraph{LLM performance.} Table~\ref{tab:f1scores} reports GPT-5.1's validation performance (F1, precision, recall) on the seven most frequent codes in the analytic sample. We also conducted an error analysis, which surfaced misclassification patterns that we attempted to address by revising code definitions in the prompts and resolving inconsistencies in the human-annotated validation set. After updating definitions, we re-ran the GPT 5.1 on the validation set. The F1 scores increased for each feature, as shown in Table~\ref{tab:f1scores}, which reports the F1 scores before and after updating definitions and human annotations. Two codes saw the largest change in F1 scores due to updating definitions, Claim (+0.104), and updating inconsistencies in human annotation, Question (+0.23). The remaining codes used for this analytic sample saw a smaller increase in F1 performance (+0.021 - +0.061).

\begin{table}[h!]
\centering
\caption{GPT-5.1 Scores by Feature Before and After Definition Update}
\label{tab:f1scores}
\begin{tabular}{l c c c c}
\toprule
\textbf{Feature} & \textbf{Old F1} & \textbf{New F1} & \textbf{New Precision} & \textbf{New Recall} \\
\midrule

 Off-Task      & 0.764 & 0.817 & 0.734 & 0.921 \\
 Understanding & 0.639 & 0.700 & 0.605 & 0.830 \\
 Recording    & 0.661 & 0.682 & 0.564 & 0.863 \\

\midrule

 Question  & 0.717 & 0.947 & 0.955 & 0.938 \\
 Claim     & 0.653 & 0.757 & 0.894 & 0.656 \\
 Next-Step & 0.586 & 0.631 & 0.515 & 0.815 \\
 Disagree  & 0.429  & 0.473 & 0.481 & 0.464 \\

\bottomrule
\end{tabular}
\end{table}

The process of creating the validation set surfaced limitations that motivate the rest of the paper. The annotation team found that disambiguating certain talk moves required context beyond text---intonation, students' intent, what they were pointing to or referring to. This was difficult even for human experts, and more so for LLMs, which lacked the annotators' background as educators and their experience listening to youth talk. In early runs of the LLM annotation, the models struggled to reach a 0.7 F1 score on most features. This gap was exacerbated by certain features being infrequent and more conceptual, requiring inference about intent rather than recognition of surface features. After adjusting definitions, the F1 scores improved but for some features---most notably Next-Step and Disagree (Table~\ref{tab:f1scores})---they remained low. This validation process motivated the need to find other ways to validate LLM-based measures of student talk and to re-contextualize the transcripts on which they depend.

\section{Ethnographically-Oriented Methods}
\label{sec:ethnographic_methods}

The epistemic shift in the validation of the LLM-based measures of student talk consists of two shifts, which include re-contextualizing student talk and bringing youth into the validation process. We employed a set of ethnographically-oriented methods as part of these shifts: participant observations, student interviews, focus groups, and member checks. 

\paragraph{Focal classroom and participants.} For the 2025--2026 case study, we selected one 8th-grade math classroom from a large urban school district on the West Coast. The classroom was selected from participating classrooms, as the school primarily serves multilingual youth from African-American and Latinx families. The classroom teacher had participated in the broader study during the prior year, but her students were new to it. From the students who consented to be recorded, we selected four focal students because they worked in the same pairs frequently, which allowed us to discuss their collaborative dynamics: Ximena and Destiny, two girls (Ximena is Latina and a designated English Learner; Destiny is Latina and Haitian), and Diego and Drake, two boys (Diego is Latino; Drake is African-American). Across the school year, I, the first author, conducted seven participant observations, four interviews, two focus groups, and four member checks with these students. We also selected two transcripts from the focal classroom as the analytic sample for the case study; the same excerpts were used in the interviews, focus groups, and member checks. The transcripts were the two most recent transcripts from lessons the students had done, so that the conversations would be more recent in the students' memories for the participant retrospection, interviews, and focus groups.

\paragraph{Positionality.} I, the first author, am an immigrant Latina woman who is multilingual and grew up in large urban city on the West Coast. I also previously taught as a K-5 classroom teacher. This background allowed me to speak in both Spanish and English with the youth, and shaped how I approached relationship-building with the teacher and students. 

\paragraph{Participant observations.} I sat with various student groups during observations to get to know the youth and notice how their group work unfolded. After each visit, I wrote field notes capturing the classroom environment, the math activities, and my own reflections on the interactions I observed.  These initial participant observations supported relationship-building between the teacher, the youth, and me, which helped create a more comfortable environment in the interviews and focus groups.

\begin{figure}[h!]
  \centering
  \includegraphics[width=.6\linewidth]{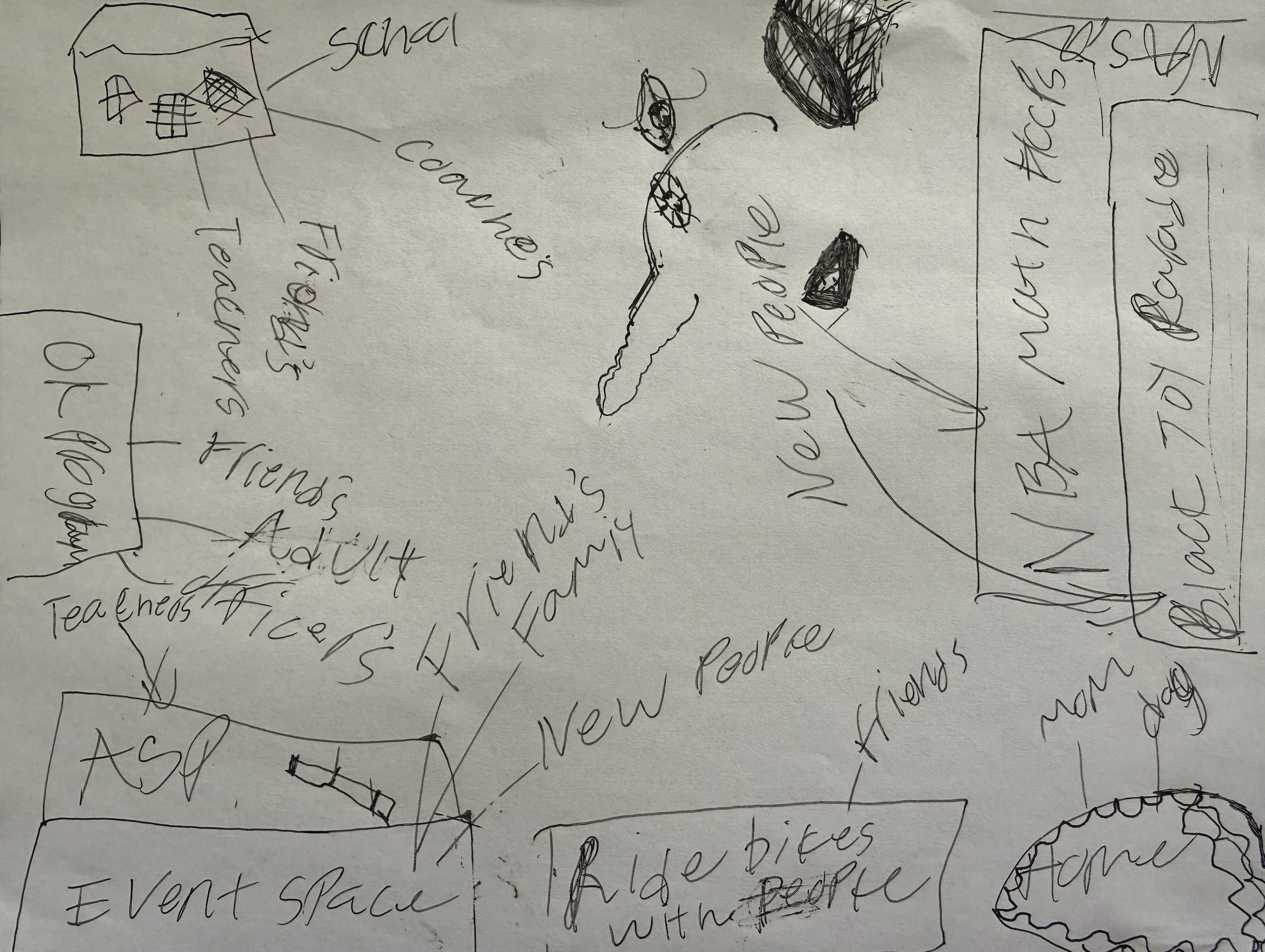}
  \caption{Example of Language Mapping Activity}
  \Description{student map}
\end{figure}
\paragraph{Student interviews.}  In one-on-one interviews, students participated in two activities: a language mapping activity and participant retrospection. Building on Martínez and Mejía, \cite{ martinez_looking_2020}, the language mapping activity asked youth to reflect on the spaces and places they traverse daily and weekly, the people they interact with, and how they use language across those audiences and interlocutors. The purpose of this activity was to build a more expansive picture of students’ linguistic repertoires both within and beyond math class. 

\paragraph{Participant retrospection and member checks.} During retrospection, youth revisited an excerpt of one of their math class conversations and reflected on what they were doing, thinking, and feeling at the time. They were then shown the LLM's annotations of that same excerpt and asked whether they agreed or disagreed with how the model had classified their talk, what they would change, and what they felt the model had missed. The full protocol is included in Table~\ref{tab:membercheck} in the Appendix.

\paragraph{Focus groups.} Focus groups paired each focal student with a peer with whom they had worked before. They revisited the same transcript excerpts used in the interviews and reflected on their group dynamics, collaboration, and what it was like to talk with one another in math class. Finally, students were asked to reflect on their group dynamics and collaboration with their classmates.

\paragraph{Analysis.} From the participant observations, I thematically grouped field notes into contextual factors that LLM-based measures miss (e.g., environmental, social, multimodal, and relational dimensions). These contextual factors, which informed how students participated, included information that was not supplied to the LLMs about the multimodal context of the classrooms and would not be represented in the transcription alone. For the member checks, I tabulated each instance where a student disagreed with an LLM classification (Table~\ref{tab:member-check-heatmap}) and sorted the disagreements into two categories: disagreement with the LLM's classification of a given utterance, or disagreement with the coding scheme. These groupings surface misalignments between how students saw their talk and how the LLMs and coding scheme classified their talk.

\section{Results}
\label{sec:results}
\subsection{Re-Contextualization: Beyond Text-Only Representations of Language}
When classroom interactions are reduced to transcripts of verbal contributions, key dimensions of participation are lost. The physical and socio-cultural environment, multimodal communication (silence, gesture, tone), teacher and student characteristics and relationships, the curriculum and lesson structure, and the broader language ecology of the classroom are missing from transcripts. Due to this, LLM-based measures that rely on such transcripts alone operate on an incomplete representation of classroom activity. 

By using ethnographically-oriented methods, we recover much of what transcripts lose by treating student talk as a situated, embodied, social, and cultural experience. In the sections that follow, we describe five dimensions of context surfaced through participant observations, interviews, and focus groups, each illustrated with examples from the focal classroom. These contextual factors allow us to (re)interpret the transcripts and create more holistic representations of student math talk and participation, and ultimately, center the student experience.

  \begin{figure}[h!]
  \centering
  \includegraphics[width=.5\linewidth]{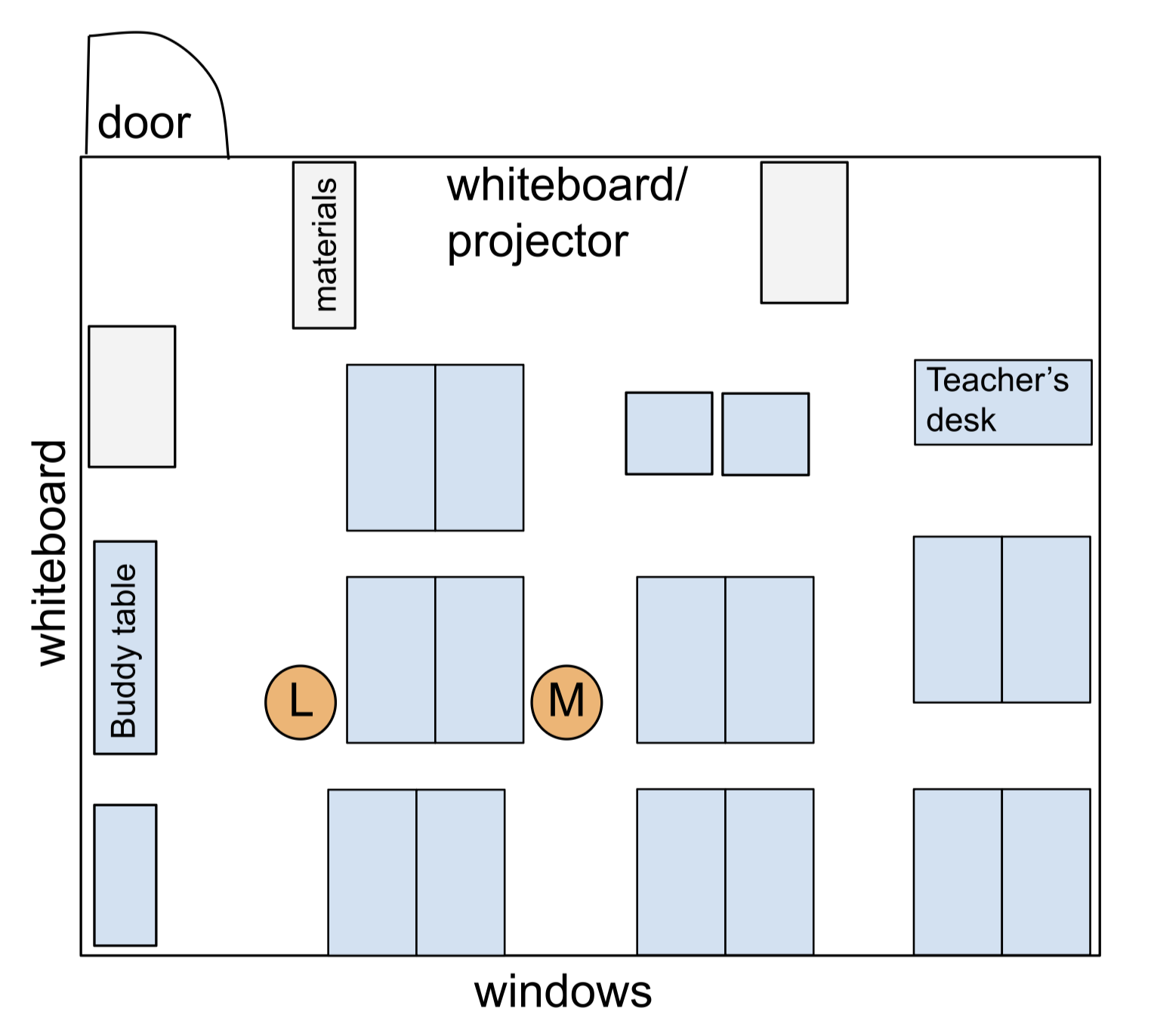}
  \caption{Layout of Ms. Richer's classroom mapping where Luna and Malik sat during a class session.
  \label{fig:layout}
  }
  \Description{Diagram of a classroom layout with desks arranged in groups. L is labeled for student Luna and M is labeled for student Malik}
\end{figure}

\subsubsection{School and Classroom Environment}
The focal school is situated in a large urban city on the West Coast in a residential neighborhood with mostly single story homes. Down the street from the school, there were businesses with signs written in Spanish and English. In the afternoons, multiple paleteros (ice cream sellers) and snack carts lined up around the front left corner of the school to sell paletas (popsicles), fruit or chicharrones de harina (wheat pinwheels) with lime and hot sauce, chips and sweets. Ms. Richer’s 8th-grade math classroom was on the first floor of a two-story building. At the time that data was collected for this study, a decoration of La Llorona (The Weeping Woman) was visible on the classroom door. Inside, the classroom was often dimly lit with shades pulled down and light streaming in from a narrow gap below the shades. Small blue string lights highlighted the contour of the whiteboard at the front of the classroom. Students were assigned seats in groups of about four per table with the only exception being the two single seat desks by the front of the classroom. The classroom was tight with only about one to two feet between any tables---Ms. Richer measured the spacing to be just right for her to squeeze through between tables. All along the walls there were posters and signs including math concepts, classroom expectations, voice level descriptors, and hand signals for the bathroom, questions, and other requests students might have. In the middle of the whiteboard at the front of the class was an agenda for the day. To its side there were class point and seating chart posters. 

The physical space in the room impacted how students worked together. In one observation, I sat with two students, Malik, an African-American boy, and Luna, a Latina girl, who were sitting across from each other. The group tables were actually two smaller rectangular tables set up next to each other to form a square (see Figure~\ref{fig:layout}). The distance between the two tables created a wide enough gap that if two students sat across from each other it was harder to hear with other conversations going on, and it made talking across the two tables less inviting. Instead of talking to one another during the pair-share time, Malik and Luna continued working in silence independently.

\subsubsection{Classroom Curriculum and Lesson Structure}

Classroom routines further structure when and how students are expected to talk. The focal classroom had structured time for when to talk and when not to talk. Ms. Richer followed a similar agenda during observations starting with a warm-up exercise, then an activity covering the current material the class was learning with designated “mild,” “medium,” and “spicy” challenge levels, and closing with an exit ticket. A typical routine included Ms. Richer starting the class with instructions and 3-5 minutes of independent work at "voice level zero," meaning there should be no talking. Then, Ms. Richer asked students to turn to their group or partner and share for 1-2 minutes at a normal talking voice level. Finally, Ms. Richer asked students to share-out in a whole class discussion for an additional 2-3 minutes. Ms. Richer typically asked students to share with a partner or share out with the class in between each activity, with lo-fi music playing in the background and a timer counting down for most activity sections. During my observations, the majority of students were quiet during the class, both in group work time and during the whole class share-out. Even during partner talk time or class share-outs, a majority of youth had their heads down looking at their worksheet continuing to work independently. Some of this was structural: parts of the curriculum directed students to complete activities individually on computers, which did not invite collaboration. For this project, we focused on student-to-student conversation, so the whole class share-outs were not part of the final transcript. 

Knowing the lesson structure matters for interpretation. Students were constrained by how much time and how many opportunities they had to talk with peers, and those constraints shaped the distribution of talk in ways that are invisible from the transcripts alone. This makes it difficult to interpret measures of participation without the context of the lesson plan.

\subsubsection{Multimodal Communication}

Two kinds of multimodal information are lost in the transcripts: non-verbal demonstrations of mathematical thinking and communication, and the communicative work done by tone, prosody, and silence.

During class, students engaged in many non-verbal actions that demonstrate their learning and mathematical thinking, which was not captured by transcribed speech. As mentioned above, during my observations, the majority of the class worked independently and quietly on math activities both during the independent work time and the pair-share time. Despite not engaging in verbal communication, most students continued to write equations and solve problems on their handouts.  

Multimodal data about student conversations can also reflect individual traits and dispositions that shape participation in ways that coding schemes and LLM-based measures do not capture. For example, Carlos, a Latino boy, rarely spoke in whole-class or group conversations but he worked diligently and quietly on his own. Drake, an African-American boy, was charismatic and outgoing, frequently contributing both in groups and during whole class share-outs. Interpreted only through what an LLM has been told counts as academic talk via the coding scheme, Carlos and Drake look like opposite ends of a participation scale. This difference between them could easily be mischaracterized as reflecting differences in engagement, whereas it may, in fact, be reflecting differences in their preferred \emph{mode} of engagement. Coding schemes built around spoken academic talk risk mischaracterizing those differences as gaps in ability or attention.
  
Multimodal data about the classroom conversations also helps when interpreting what students meant in the context of their conversations. Tone, silence, prosody, and body movements all help make meaning and these communicative dimensions are lost in transcription. Within transcribed speech, attending to tone or prosody can disambiguate a student's intent. For example, when "Yeah" is said in a transcript, it can be a signal of agreement or a filler word, and the disambiguating signal is in the voice, not the transcript. Gesture and posture similarly shape what an utterance means in context yet remain unavailable to text-based LLM measures.

\subsubsection{Language Ideologies and Norms}
Institutional language ideologies and norms also shape student discourse. Although many students in the classroom spoke Spanish as a home language, instruction was conducted in English, and the use of other languages was not explicitly encouraged. As a result, transcripts may not capture the broader linguistic repertoires students draw on in other contexts or outside of recording times. Ms. Richer is a white woman in her mid 30s who had been teaching math for 10 years at the time of this study. She speaks English and shared that she understands “some” Spanish. Ms. Richer explained that while the school did not explicitly prohibit using other named languages like Spanish, they did not explicitly encourage it in the classroom either. Most class discussion used English, with some colloquial language from Ms. Richer. The class was majority Latine and bilingual in Spanish and English; most of those students were Reclassified, with a handful designated English Learners. The students designated English-Only were primarily African-American. This was reflective of the school's broader demographics, where the majority of students were Latine or African-American. 

Students' use of home languages was discussed in student interviews and the language mapping activity. For example, Ximena, who is a designated English Learner, shared that even though she speaks Spanish, she would only use it in certain contexts at school, such as during Physical Education class, where she helped translate for Spanish-speaking peers. However, in math class, Ximena shared that she did not use Spanish. During my observations, I only heard Ximena use Spanish a handful of times. Ximena's classmate and friend, Destiny, also shared that she knew some Spanish and Haitian Creole, but did not use these at school. Both students also used features of African-American English though neither named this language variety in their language mapping activity. When I asked the students about whether they thought they used African-American English, Ximena looked over at Destiny and said, "No, I thought I was just talkin' normal English." Destiny smiled and added, "Right, like I never noticed anything different." Looking only at LLM-based measures of student talk based on transcripts would fail to capture the expansiveness of students' linguistic repertoires, the context for why they may change the way they speak with different audiences and in different places, and the way their talk is informed and constrained by language ideologies.

\subsubsection{Relationships}
In focus groups, students emphasized that interpersonal relationships affected how comfortable they felt talking to their peers in class. In response to a question about what it was like to be paired with Ximena, Destiny shared: “I feel like sitting next to her, I just feel comfortable.” Elaborating on whether it was important to feel comfortable with her peer, Destiny said, "Yes because I’m a really anxious person, like I have lots of anxiety and like social anxiety, and so like being around people that I’m comfortable with instead of sitting next to somebody I don’t know is a lot easier for me.” These relationships are often not explicit in recordings and are not captured in LLM-based coding schemes. Students are aware that their experience collaborating with peers may be different based on preexisting relationships. 

Students described their relationship with their teacher as similarly important. In student interviews and focus groups, students expressed feeling more comfortable with Ms. Richer than with other teachers at their school. They reported that this impacted how they talked in the classroom, including their willingness to engage in off-task talk. Ximena shared in an interview that Ms. Richer was "more calmer and that’s why everybody’s her favorite teacher". Ms. Richer's relationship with her students and her disciplinary choices made Ximena feel more comfortable talking in her class and made Ms. Richer her favorite teacher. For Ximena and Destiny, that comfort shaped what they felt able to say. Such dynamics are not captured in LLM-based coding schemes, but they help explain variation in participation that transcript-based measures would otherwise obscure.

\subsection{"ChatPT [sic] needs to get his life together": Students Engaging with LLM Based Talk Measures}

Re-contextualizing transcripts goes some distance toward centering student experience, but it does not fully bring students into the validation process. Students are experts on their own talk and bringing them into the validation process can provide nuance, depth, and contrasting viewpoints that adult annotators, researchers, and LLMs cannot provide alone. Student member checks revealed that discrepancies between LLM classifications and student interpretations were not limited to isolated errors. They reflected deeper misalignments between how coding schemes define academic talk and how students understand their own participation. While students generally agreed with the LLM classifications, the moments of contestation---and the reasons they gave for them---highlighted misalignments with the LLMs' classification of a given utterance, and with the coding scheme itself. Table~\ref{tab:member-check-heatmap} shows the highest-frequency disagreements during member checks. Students were especially vocal about the Claim and Off-task codes and pushed back on these labels by foregrounding the intentions behind their utterances.

\begin{table*}[h]
\centering
\caption{Disagreement Rates During Student Member Checks for Analytic Codes}
\label{tab:member-check-heatmap}

\begin{tabular}{lrrrrrrr}
\toprule
\textbf{Student} & \textbf{Off-Task} & \textbf{Understanding} & \textbf{Recording} & \textbf{Question} & \textbf{Claim} & \textbf{Next-Step} & \textbf{Disagree} \\
\midrule
Diego   & 0/2 & 0/6 & -- & 0/10 & \cellcolor{gray!35}5/12 & 0/1 & \cellcolor{gray!80}1/1 \\

Drake   & \cellcolor{gray!25}1/5 & \cellcolor{gray!20}1/4 & -- & 0/6 & 0/6 & \cellcolor{gray!80}1/1 & 0/1 \\

Destiny & 0/8 & 0/5 & \cellcolor{gray!80}1/2 & 0/8 & \cellcolor{gray!20}1/4 & \cellcolor{gray!30}1/3 & 0/2 \\

Ximena  & \cellcolor{gray!50}5/8 & 0/4 & -- & \cellcolor{gray!20}1/4 & \cellcolor{gray!20}3/17 & 0/1 & -- \\
\bottomrule
\end{tabular}

\vspace{2pt}

\parbox{\textwidth}{
\small
\textit{Note.} Cells show the disagreement counts divided by the total count
for each code. Dashes indicate that the code did not occur for that student
in the analytic excerpt.
}

\end{table*}

Students argued for more expansive definitions of what counted as math talk. Ximena pushed back on the Off-task label three times because these were moments when she explicitly said "math" and related to her math learning experiences. In the first excerpt (Table~\ref{tab:excerpt1}), where Ximena and Destiny discussed how to solve a problem, the LLM annotated lines 8-9 (``It's why we don't do math. Math pisses me off.'') as off-task. From Ximena's perspective, expressing frustration was part of working on the math task.  The pattern repeated in a second excerpt (Table~\ref{tab:excerpt2}), where Destiny checks her answer with Ximena and Ximena replies, ``I don't know see honey bunch, you're asking the wrong person over here. Let's be for real. you're asking the wrong person.'' The LLM marked both lines as off-task. Ximena argued they were not, since she was ``still talking about the work.'' A third instance occurred mid-conversation about a math problem, when Ximena said, "[Teacher], I think I'm actually smart. I don't even know why I'm in eighth grade. I'm actually supposed to be in high school." Her position was consistent: "I think it was on task because I was talking about math." In all three of these instances, Ximena engaged actively with math tasks and made comments about her personal math experience. Ximena was quick to say, "ChatPT [sic] needs to get his life together," in response to how the LLM was annotating her talk. Not only was Ximena offering a more expansive idea of what could be considered math talk, she was also pointing out the LLM's role in this process. By engaging Ximena in the conversation about her own talk, the epistemic authority shifted to include her as a part of the knowledge production process.

\begin{table*}[h!]
\centering
\caption{Excerpt 2 from Classroom Conversation between Ximena and Destiny}
\label{tab:excerpt2}
\begin{tabular}{p{0.12\textwidth} p{0.08\textwidth} p{0.72\textwidth}}
\toprule
\textbf{Speaker} & \textbf{Line} & \textbf{Dialogue} \\
\midrule
Destiny  & 1 & ``Divided by six.'' \\
 & 2 & ``Negative one. Right? Am I, am I dumb or am I dumb?'' \\
Ximena  & 3 & ``I don't know see honey bunch, you're asking the wrong person over here.'' \\
 & 4 & ``Let's be for real. you're asking the wrong person.'' \\

\bottomrule
\end{tabular}
\end{table*}

Students also added more clarity around their intentions and what was missed by the coding scheme used for LLM classification. Ximena and Diego had similar utterances that were coded by the LLM as Claim, but they disagreed with this code. In a recording, Ximena had said, "Ten, twenty, thirty, forty," and reflecting on this she thought this should have been marked as showing reasoning: "I say reasoning...because I was trying to figure out why it's this number, so I was counting." Diego had an instance where he was also counting out loud: "Sixty. Seventy." and similarly disagreed with the Claim label: "Nah I was just like counting...I think it's different because a claim.. you're saying that's the answer or something." For both students, the intention behind their utterance was different from a claim. As Diego explained, a claim involves asserting a final answer, whereas counting reflects an intermediate step. This distinction, while captured by the coding scheme, was not replicated by the LLM-based measures, which classify utterances based on the raw data rather than intended function.

Reflecting on their original intentions in conversations, students described possible new codes for their talk that were missing from the coding scheme. Diego had another instance where he was talking to Drake about a possible solution and said: "I feel like it's twenty-eight for some reason." The LLM classified this as a Claim and Disagree. Diego clarified, "No, it was just like a guess, I was just guessing I wasn't claiming that it was.. wasn't denying it wasn't also 28." In this moment, Diego was able to make a guess about a possible solution and hold multiple possibilities in his head. When I asked how he would categorize this instead, Diego asked, "There's no guess?" There were no codes in the coding scheme for brainstorming or guessing, so this function was missed.

Member checks reveal that coding schemes embed assumptions about what counts as meaningful participation, and that those assumptions do not always align with how students themselves understand their talk. Centering student interpretations can both help correct misalignments and surface categories the scheme failed to anticipate.

\section{Discussion}

Our findings highlight two epistemic shifts in the design of LLM-based measures of student talk. The first concerns the data: by reducing rich, embodied classroom interactions to text and then to discrete labels, LLM-based systems omit relational, spatial, socio-cultural, and multimodal dimensions of participation by design. These omissions are not errors that can be corrected through improved prompting or larger datasets; they are structural limitations of applying text-based classification to complex social interaction \cite{yeh_beyond_2025, stewart_beyond_2026}.  Because they take conversations out of context and exclude student input, these omissions also de-center the youth whose talk is being measured. 

The second concerns validation. Standard validation practices ask whether a model reproduces expert labels, but validation is itself a knowledge-production practice: it determines whose interpretation of the data counts. By involving youth in validation, we shift epistemic authority over interpretation from researchers and models to students themselves (Figure 1). This re-centering of youth allows us to better understand not only how students see their own math talk, but also whether these talk measures are meaningful to them. Asking for the perspective of students allows us to validate if the talk measures are accurate to their original intentions and, beyond this, gives insight as to what youth see as part of the math learning experience. The disagreements the students in this study raised were not simply classification errors, but alternative interpretations of what their talk meant. In other words, student were not simply interested in using the available codes provided by the research team to improve model performance, they wanted to change the interpretations of existing codes, and add codes they felt were missing. When Ximena contested her talk being labeled as off-task, for instance, she was asserting that her math conversations include social and affective dimensions---an assertion that aligns with prior work showing that off-task talk can serve a range of social and communicative purposes \cite{langer-osuna_exploring_2020, martinez_puras_2014}. Students also added nuance to their original intentions, which highlighted misalignments with the coding scheme. These misalignments might cause LLMs to misattribute talk moves and miss significant contributions. Incorporating student perspectives is, therefore, not just a methodological choice, but a necessary step toward creating more equitable and representative forms of AI in education.

Taken together, these findings argue for participatory validation as a core component of responsible AI in education. Tools and research that measure student discourse can shape how participation, collaboration, and learning are understood and evaluated. Without incorporating the perspectives of the communities being measured, these outcomes risk reinforcing narrow or misaligned definitions of meaningful participation, particularly for multilingual and marginalized students. Centering youth as epistemic authorities additionally improves the creation and interpretation of measures by grounding these measures in the lived experiences of the youth. We recommend that scholars using LLM-based discourse measures triangulate data from multiple methods and seek input from participating communities to validate and improve measures. This is especially important for combating the biases and harms that LLMs can reproduce and that disproportionately affect marginalized youth.

\section{Limitations and Ethical Considerations}

Several limitations bound the claims of this paper. First, our framing intentionally treats student learning as more than spoken contributions, and we address this through methodological triangulation. But the same methods that allow us to re-contextualize a small number of conversations also constrain how many we can analyze. Part of the work in the focal classroom included building relationships slowly and intentionally over the course of months, which limits the total number of students and classrooms under consideration. The tradeoff is that we cannot generalize from one classroom and four students. A larger or more varied sample would likely surface participation patterns and disagreements we did not observe.

Second, our participatory engagement was scoped to the validation phase. Youth provided feedback on the LLM's outputs and on the coding scheme, but they were not involved in designing the coding scheme or the LLM prompts in the first place. A fuller participatory approach, in which youth shape the underlying definitions of math talk before classification begins, would likely surface further mismatches that our member checks missed.

Finally, the use of LLM-based measures raises concerns not only about the bias and harms that LLMs have been shown to cause for users, but also about the environment and communities near data centers. This is especially concerning for the larger models that cannot be locally run. Our team conducted two annotation runs with GPT 5.0 via API calls, one for the full set of features, one for a reduced set of features with updated definitions, and one for the two analytic excerpts. While our validation set and analytic sample is small, the impact of our research must also be taken into account as we look towards future work. As we center equity in our work, we must also account for the impact of our research on communities beyond our focal classroom, including those most affected by the infrastructure that LLMs depend on.

\section{Future work}

The limitations of this work point to natural extensions: a broader sample, fuller co-design and participatory approaches with youth, and smaller, locally-runnable models. Beyond these, our findings open several further directions for inquiry. First, extending this work beyond mathematics classrooms, into after-school programs, extended-day spaces, or other subject-area classrooms, would help us understand how youth language practices and the categories that describe them shift across settings. Second, future research could examine what happens downstream of measure development. When teachers, students, or administrators see LLM-based discourse outputs, how do those outputs shape participation, instruction, and student-teacher relationships? Finally, the participatory validation approach we describe could be applied to other AI tools used in educational settings, such as automated feedback systems, assessment tools, or behavior detection, to test whether the misalignments we observed are specific to discourse measurement or reflect broader gaps between AI-generated representations and the experiences of the students they describe.

\section{Conclusion}
This paper argues that standard validation practices for LLM-based measures of classroom discourse (i.e., expert annotation, F1 scores, held-out test sets) are insufficient for capturing the full meaning of student participation and ensuring these measures are meaningful and equitable. Through a case study of multilingual youth in one 8th-grade math classroom, we showed that ethnographically-oriented methods recover dimensions of classroom interaction that text-only transcripts cannot represent, and that students themselves contest both model classifications and the coding schemes used to produce them. Treating youth as epistemic authorities on their own talk reframes validation as a knowledge-production practice in which the people being measured help decide what the measurements mean. As LLM-based tools become more widely used in educational contexts, participatory and community-engaged approaches that center the voices of students are essential for ensuring that these systems are both accurate and equitable. Students are experts with respect to their own experiences, and epistemic authority should be shared with them when evaluating and designing tools to interpret their talk.

\begin{acks}
I would like to thank my husband, Reydrick Santos-Deonizio, and family for their support during the writing of this paper. We thank all the participants in the study and the students and teacher in the focal classroom. Thank you to Ms. Richer's class for welcoming us into your community and letting us observe the brilliant ways you all engage in math. We also thank Helen Higgins who helped set up the study infrastructure, Lucía Langlois who helped with editing, Dr. Eujin Park who provided thoughtful feedback, and Dr. Christina Krist who supported through thought partnering. This work was funded by the Bill and Melinda Gates Foundation. 
\end{acks}

\bibliographystyle{ACM-Reference-Format}
\bibliography{sample-base}

@inproceedings{navarro_re-imagining_2022,
	address = {Taiwan},
	series = {{CSCW} {Companion} '22},
	title = {Re-imagining systems in the realm of immigration in higher education through participatory design},
	isbn = {978-1-4503-9190-0},
	url = {https://dl.acm.org/doi/10.1145/3500868.3559457},
	language = {en},
	urldate = {2026-08-05},
	booktitle = {Companion {Publication} of the 2022 {Conference} on {Computer}-{Supported} {Cooperative} {Work} and {Social} {Computing}},
	publisher = {Association for Computing Machinery},
	author = {Navarro, Maria Conchita A. and Shaer, Orit},
	month = nov,
	year = {2022},
	pages = {76--79},
}

@inproceedings{gak_reflecting_2024,
	address = {New York, NY, USA},
	series = {{CSCW} {Companion} '24},
	title = {Reflecting on the relational: {Youth}-centered approaches to living with technology},
	isbn = {979-8-4007-1114-5},
	shorttitle = {Reflecting on the {Relational}},
	url = {https://dl.acm.org/doi/10.1145/3678884.3682047},
	doi = {10.1145/3678884.3682047},
	urldate = {2026-08-04},
	booktitle = {Companion {Publication} of the 2024 {Conference} on {Computer}-{Supported} {Cooperative} {Work} and {Social} {Computing}},
	publisher = {Association for Computing Machinery},
	author = {Gak, Liza},
	month = nov,
	year = {2024},
	pages = {27--30},
}

@inproceedings{ajmani_power_2025,
	address = {New York, NY, USA},
	series = {{CSCW} {Companion} '25},
	title = {Power, participation, and knowledge production: Technology mediated epistemic (in)justice},
	isbn = {979-8-4007-1480-1},
	shorttitle = {Power, {Participation}, and {Knowledge} {Production}},
	url = {https://dl.acm.org/doi/10.1145/3715070.3747343},
	doi = {10.1145/3715070.3747343},
	urldate = {2026-08-03},
	booktitle = {Companion {Publication} of the 2025 {Conference} on {Computer}-{Supported} {Cooperative} {Work} and {Social} {Computing}},
	publisher = {Association for Computing Machinery},
	author = {Ajmani, Leah Hope},
	month = oct,
	year = {2025},
	pages = {51--53},
}

@inproceedings{stewart_beyond_2026,
	address = {New York, NY, USA},
	series = {{LAK} '26},
	title = {Beyond the numbers: {Socio}-{Cultural} context as a frame for learning analytics},
	isbn = {979-8-4007-2066-6},
	shorttitle = {Beyond the {Numbers}},
	url = {https://dl.acm.org/doi/10.1145/3785022.3785076},
	doi = {10.1145/3785022.3785076},
	urldate = {2026-05-05},
	booktitle = {Proceedings of the {LAK26}: 16th {International} {Learning} {Analytics} and {Knowledge} {Conference}},
	publisher = {Association for Computing Machinery},
	author = {Stewart, Angela E.B. and Hutt, Stephen},
	month = apr,
	year = {2026},
	pages = {852--858},
}

@article{esmonde_power_2013,
	chapter = {Journal for Research in Mathematics Education},
	title = {Power in numbers: {Student} participation in mathematical discussions in heterogeneous spaces},
	volume = {44},
	issn = {0021-8251, 1945-2306},
	shorttitle = {Power in {Numbers}},
	url = {https://pubs.nctm.org/view/journals/jrme/44/1/article-p288.xml},
	doi = {10.5951/jresematheduc.44.1.0288},
	language = {en\_US},
	number = {1},
	urldate = {2026-05-02},
	journal = {Journal for Research in Mathematics Education},
	publisher = {National Council of Teachers of Mathematics},
	author = {Esmonde, Indigo and Langer-Osuna, Jennifer M.},
	month = jan,
	year = {2013},
	pages = {288--315},
}

@article{langer-osuna_so_2020,
	title = {“{So} what are we working on?”: how student authority relations shift during collaborative mathematics activity},
	volume = {104},
	issn = {1573-0816},
	shorttitle = {“{So} what are we working on?},
	url = {https://doi.org/10.1007/s10649-020-09962-3},
	doi = {10.1007/s10649-020-09962-3},
	language = {en},
	number = {3},
	urldate = {2026-05-01},
	journal = {Educational Studies in Mathematics},
	author = {Langer-Osuna, Jennifer and Munson, Jen and Gargroetzi, Emma and Williams, Immanuel and Chavez, Rosa},
	month = jul,
	year = {2020},
	pages = {333--349},
}

@article{vakil_youth_2022,
	title = {Youth as philosophers of technology},
	volume = {29},
	issn = {1074-9039},
	url = {https://doi.org/10.1080/10749039.2022.2066134},
	doi = {10.1080/10749039.2022.2066134},
	number = {4},
	urldate = {2026-05-01},
	journal = {Mind, Culture, and Activity},
	publisher = {Routledge},
	author = {Vakil, Sepehr and McKinney de Royston, Maxine},
	month = oct,
	year = {2022},
	note = {\_eprint: https://doi.org/10.1080/10749039.2022.2066134},
	pages = {336--355},
}

@article{martinez_black_2022,
	title = {Black lives matter versus {Castañeda} v. {Pickard}: a utopian vision of who counts as bilingual (and who matters in bilingual education)},
	volume = {21},
	issn = {1573-1863},
	shorttitle = {Black lives matter versus {Castañeda} v. {Pickard}},
	url = {https://doi.org/10.1007/s10993-021-09610-3},
	doi = {10.1007/s10993-021-09610-3},
	language = {en},
	number = {3},
	urldate = {2026-05-01},
	journal = {Language Policy},
	author = {Martínez, Ramón Antonio and Martinez, Danny C. and Morales, P. Zitlali},
	month = sep,
	year = {2022},
	pages = {427--449},
}

@article{hill_language_1998,
	title = {Language, race, and white public space},
	volume = {100},
	issn = {0002-7294, 1548-1433},
	url = {https://anthrosource.onlinelibrary.wiley.com/doi/10.1525/aa.1998.100.3.680},
	doi = {10.1525/aa.1998.100.3.680},
	language = {en},
	number = {3},
	urldate = {2026-05-01},
	journal = {American Anthropologist},
	author = {Hill, Jane H.},
	month = sep,
	year = {1998},
	pages = {680--689},
}

@article{drageset_different_2015,
	title = {Different types of student comments in the mathematics classroom},
	volume = {38},
	issn = {0732-3123},
	url = {https://www.sciencedirect.com/science/article/pii/S0732312315000152},
	doi = {10.1016/j.jmathb.2015.01.003},
	urldate = {2026-05-01},
	journal = {The Journal of Mathematical Behavior},
	author = {Drageset, Ove Gunnar},
	month = jun,
	year = {2015},
	pages = {29--40},
}

@article{baze_call_2025,
	title = {A call to explicitly name and account for power in epistemic agency research},
	volume = {109},
	issn = {1098-237X},
	url = {https://onlinelibrary.wiley.com/doi/10.1002/sce.21966},
	doi = {10.1002/sce.21966},
	language = {en},
	number = {5},
	urldate = {2026-05-01},
	journal = {Science Education},
	publisher = {John Wiley \& Sons, Ltd},
	author = {Baze, Christina and González-Howard, María},
	month = sep,
	year = {2025},
	pages = {1499--1505},
}

@article{mathayas_re-indexing_2026,
	title = {Re-{Indexing} epistemic responsibility: {A} grammatical analysis of how a teacher made space for students' epistemic agency},
	volume = {63},
	copyright = {© 2025 The Author(s). Journal of Research in Science Teaching published by Wiley Periodicals LLC on behalf of National Association for Research in Science Teaching.},
	issn = {1098-2736},
	shorttitle = {Re-{Indexing} {Epistemic} {Responsibility}},
	url = {https://onlinelibrary.wiley.com/doi/abs/10.1002/tea.70023},
	doi = {10.1002/tea.70023},
	language = {en},
	number = {1},
	urldate = {2026-05-01},
	journal = {Journal of Research in Science Teaching},
	author = {Mathayas, Nitasha and Krist, Christina},
	year = {2026},
	note = {\_eprint: https://onlinelibrary.wiley.com/doi/pdf/10.1002/tea.70023},
	pages = {62--82},
}

@article{yeh_beyond_2025,
	title = {Beyond verbal: {A} methodological approach to highlighting students’ embodied participation in mathematics classroom},
	volume = {54},
	issn = {0013-189X},
	shorttitle = {Beyond {Verbal}},
	url = {https://journals.sagepub.com/doi/full/10.3102/0013189X241310169},
	doi = {10.3102/0013189X241310169},
	number = {2},
	urldate = {2026-04-24},
	journal = {Educational Researcher},
	publisher = {American Educational Research Association},
	author = {Yeh, Cathery and Reinholz, Daniel Lee and Lee, Hakeoung Hannah and Moschetti, Mariah},
	month = mar,
	year = {2025},
	pages = {103--110},
}

@article{gutierrez_at_2006,
	title = {{AT} {LAST}: {The} "problem" of english learners: constructing genres of difference},
	volume = {40},
	issn = {0034-527X, 1943-2348},
	shorttitle = {{AT} {LAST}},
	url = {https://publicationsncte.org/content/journals/10.58680/rte20065110},
	doi = {10.58680/rte20065110},
	language = {en},
	number = {4},
	urldate = {2026-04-24},
	journal = {Research in the Teaching of English},
	publisher = {ncte.org},
	author = {Gutiérrez, Kris D. and Orellana, Marjorie Faulstich},
	month = may,
	year = {2006},
	pages = {502--507},
}

@article{martinez_looking_2020,
	title = {Looking closely and listening carefully: {A} sociocultural approach to understanding the complexity of {Latina}/o/x students’ everyday language},
	volume = {59},
	issn = {0040-5841},
	shorttitle = {Looking closely and listening carefully},
	url = {https://doi.org/10.1080/00405841.2019.1665414},
	doi = {10.1080/00405841.2019.1665414},
	number = {1},
	urldate = {2026-04-24},
	journal = {Theory Into Practice},
	publisher = {Routledge},
	author = {Martínez, Ramón Antonio and Mejía, Alexander Feliciano},
	month = jan,
	year = {2020},
	note = {\_eprint: https://doi.org/10.1080/00405841.2019.1665414},
	pages = {53--63},
}

@inproceedings{demszky-hill-2023-ncte,
    title = "The {NCTE} transcripts: A dataset of elementary math classroom transcripts",
    author = "Demszky, Dorottya  and
      Hill, Heather",
    editor = {Kochmar, Ekaterina  and
      Burstein, Jill  and
      Horbach, Andrea  and
      Laarmann-Quante, Ronja  and
      Madnani, Nitin  and
      Tack, Ana{\"i}s  and
      Yaneva, Victoria  and
      Yuan, Zheng  and
      Zesch, Torsten},
    booktitle = "Proceedings of the 18th Workshop on Innovative Use of NLP for Building Educational Applications (BEA 2023)",
    month = jul,
    year = "2023",
    address = "Toronto, Canada",
    publisher = "Association for Computational Linguistics",
    url = "https://aclanthology.org/2023.bea-1.44/",
    doi = "10.18653/v1/2023.bea-1.44",
    pages = "528--538"
}

@inproceedings{pugh_speech-based_2022,
	address = {New York, NY, USA},
	series = {{LAK22}},
	title = {Do speech-based collaboration analytics generalize across task contexts?},
	isbn = {978-1-4503-9573-1},
	url = {https://doi.org/10.1145/3506860.3506894},
	doi = {10.1145/3506860.3506894},
	booktitle = {{LAK22}: 12th {International} {Learning} {Analytics} and {Knowledge} {Conference}},
	publisher = {Association for Computing Machinery},
	author = {Pugh, Samuel L. and Rao, Arjun and Stewart, Angela E.B. and D'Mello, Sidney K.},
	year = {2022},
	pages = {208--218},
}

@inproceedings{park_understanding_2025,
	address = {New York, NY, USA},
	series = {{LAK} '25},
	title = {Understanding collaborative learning processes and outcomes through student discourse dynamics},
	isbn = {979-8-4007-0701-8},
	url = {https://doi.org/10.1145/3706468.3706547},
	doi = {10.1145/3706468.3706547},
	booktitle = {Proceedings of the 15th {International} {Learning} {Analytics} and {Knowledge} {Conference}},
	publisher = {Association for Computing Machinery},
	author = {Park, Seehee and Nixon, Nia and D'Mello, Sidney and Shariff, Danielle and Choi, Jaeyoon},
	year = {2025},
	pages = {938--943},
}

@article{meyer_using_2024,
	title = {Using {LLMs} to bring evidence-based feedback into the classroom: {AI}-generated feedback increases secondary students’ text revision, motivation, and positive emotions},
	volume = {6},
	issn = {2666-920X},
	shorttitle = {Using {LLMs} to bring evidence-based feedback into the classroom},
	url = {https://www.sciencedirect.com/science/article/pii/S2666920X23000784},
	doi = {10.1016/j.caeai.2023.100199},
	urldate = {2026-04-17},
	journal = {Computers and Education: Artificial Intelligence},
	author = {Meyer, Jennifer and Jansen, Thorben and Schiller, Ronja and Liebenow, Lucas W. and Steinbach, Marlene and Horbach, Andrea and Fleckenstein, Johanna},
	month = jun,
	year = {2024},
	pages = {100199},
}

@article{krist_teacher_2023,
	title = {Teacher noticing for supporting students' epistemic agency in science sensemaking discussions},
	volume = {34},
	issn = {1046-560X},
	url = {https://doi.org/10.1080/1046560X.2022.2155355},
	doi = {10.1080/1046560X.2022.2155355},
	number = {8},
	urldate = {2024-07-13},
	journal = {Journal of Science Teacher Education},
	publisher = {Routledge},
	author = {Krist, Christina (Stina) and Machaka, Nessrine and Voss, Dan and Mathayas, Nitasha and Kelly, Susan and Shim, Soo-Yean},
	month = nov,
	year = {2023},
	note = {\_eprint: https://doi.org/10.1080/1046560X.2022.2155355},
	pages = {799--819},
}

@article{pugh_say_2021,
	title = {Say what? {Automatic} modeling of collaborative problem solving skills from student speech in the wild},
	language = {en},
	author = {Pugh, Samuel L and Subburaj, Shree Krishna and Rao, Arjun Ramesh and Stewart, Angela E B and Andrews-Todd, Jessica and D’Mello, Sidney K},
	year = {2021},
}

@inproceedings{harvey_dont_2025,
	address = {New York, NY, USA},
	series = {{CHI} '25},
	title = {"{Don}'t forget the teachers": {Towards} an educator-centered understanding of harms from large language models in education},
	isbn = {979-8-4007-1394-1},
	shorttitle = {"{Don}'t {Forget} the {Teachers}"},
	url = {https://dl.acm.org/doi/10.1145/3706598.3713210},
	doi = {10.1145/3706598.3713210},
	urldate = {2026-04-09},
	booktitle = {Proceedings of the 2025 {CHI} {Conference} on {Human} {Factors} in {Computing} {Systems}},
	publisher = {Association for Computing Machinery},
	author = {Harvey, Emma and Koenecke, Allison and Kizilcec, Rene F.},
	month = apr,
	year = {2025},
	pages = {1--19},
}

@misc{ahtisham_optimizing_2026,
	title = {Optimizing {LLM} annotation of classroom discourse through multi-agent orchestration},
	url = {http://arxiv.org/abs/2603.13353},
	doi = {10.48550/arXiv.2603.13353},
	urldate = {2026-04-17},
	publisher = {arXiv},
	author = {Ahtisham, Bakhtawar and Vanacore, Kirk and Kizilcec, Rene F.},
	month = mar,
	year = {2026},
	note = {arXiv:2603.13353 [cs]},
}

@inproceedings{reitman_multi-theoretic_2023,
	address = {Cham},
	title = {A multi-theoretic analysis of collaborative discourse: {A} step towards {AI}-facilitated student collaborations},
	isbn = {978-3-031-36272-9},
	shorttitle = {A {Multi}-theoretic {Analysis} of {Collaborative} {Discourse}},
	doi = {10.1007/978-3-031-36272-9_47},
	language = {en},
	booktitle = {Artificial {Intelligence} in {Education}},
	publisher = {Springer Nature Switzerland},
	author = {Reitman, Jason G. and Clevenger, Charis and Beck-White, Quinton and Howard, Amanda and Rose, Sierra and Elick, Jacob and Harris, Julianna and Foltz, Peter and D’Mello, Sidney K.},
	editor = {Wang, Ning and Rebolledo-Mendez, Genaro and Matsuda, Noboru and Santos, Olga C. and Dimitrova, Vania},
	year = {2023},
	pages = {577--589},
}

@article{anderson_exploring_2025,
	title = {Exploring {GenAI} technologies within collaborative learning},
	url = {https://repository.isls.org//handle/1/11900},
	urldate = {2026-04-22},
	publisher = {International Society of the Learning Sciences},
	author = {Anderson, Emma and Lin, Grace C. and Farid, Amelia and Fenech, Mic and Hanks, Brandon and Klopfer, Eric and Doherty, Emily and Hirshfield, Leanne and Ko, Mon-Lin Monica and Foltz, Peter and Nguyen, Ha and Nguyen, Victoria and Ludovise, Sara and Santagata, Rossella and Cao, Lydia and Scardamalia, Marlene and Soliman, Dina and Resendes, Monica and Khanlari, Ahmad and Costa, Stacy and Chan, Carol K. K. and Nguyen, Andy},
	year = {2025},
}

@book{gonzalez_funds_2009,
	title = {Funds of knowledge: {Theorizing} practices in households, communities, and classrooms},
	publisher = {Routledge},
	author = {Gonzalez, Norma and Moll, Luis C. and Amanti, Cathy},
	year = {2009},
}

@incollection{flores_producing_2024,
	title = {Producing deficiency and erasing colonialism in the bilingual education act},
	isbn = {978-0-19-751681-2},
	url = {https://doi.org/10.1093/oso/9780197516812.003.0005},
	booktitle = {Becoming the {System}: {A} {Raciolinguistic} {Genealogy} of {Bilingual} {Education} in the {Post}-{Civil} {Rights} {Era}},
	publisher = {Oxford University Press},
	author = {Flores, Nelson},
	year = {2024},
	note = {Type: 10.1093/oso/9780197516812.003.0005},
}

@article{langer-osuna_exploring_2020,
	title = {Exploring the role of off-task activity on students’ collaborative dynamics.},
	volume = {112},
	issn = {1939-2176, 0022-0663},
	url = {https://doi.apa.org/doi/10.1037/edu0000464},
	doi = {10.1037/edu0000464},
	language = {en},
	number = {3},
	urldate = {2025-06-01},
	journal = {Journal of Educational Psychology},
	author = {Langer-Osuna, Jennifer M. and Gargroetzi, Emma and Munson, Jen and Chavez, Rosa},
	month = apr,
	year = {2020},
	pages = {514--532},
}

@article{wei_not_2022,
	title = {Not a first language but one repertoire: {Translanguaging} as a decolonizing project},
	volume = {53},
	issn = {0033-6882},
	shorttitle = {Not a {First} {Language} but {One} {Repertoire}},
	url = {https://journals.sagepub.com/doi/10.1177/00336882221092841},
	doi = {10.1177/00336882221092841},
	number = {2},
	urldate = {2025-06-01},
	journal = {RELC Journal},
	publisher = {SAGE Publications Ltd},
	author = {Wei, Li and García, Ofelia},
	month = aug,
	year = {2022},
	pages = {313--324},
}

@article{martinez_reading_2013,
	title = {Reading the world in {Spanglish}: {Hybrid} language practices and ideological contestation in a sixth-grade {English} language arts classroom},
	volume = {24},
	issn = {0898-5898},
	url = {https://www.sciencedirect.com/science/article/pii/S0898589813000235},
	doi = {https://doi.org/10.1016/j.linged.2013.03.007},
	number = {3},
	journal = {Linguistics and Education},
	author = {Martínez, Ramón Antonio},
	year = {2013},
	pages = {276--288},
}

@book{garcia_translanguaging_2014,
	address = {London},
	title = {Translanguaging},
	copyright = {http://www.springer.com/tdm},
	isbn = {978-1-349-48138-5 978-1-137-38576-5},
	url = {http://link.springer.com/10.1057/9781137385765},
	doi = {10.1057/9781137385765},
	language = {en},
	urldate = {2026-04-18},
	publisher = {Palgrave Macmillan UK},
	author = {García, Ofelia and Wei, Li},
	year = {2014},
}

@article{flores_undoing_2015,
	title = {Undoing appropriateness: {Raciolinguistic} ideologies and language diversity in education},
	volume = {85},
	copyright = {Copyright © by the President and Fellows of Harvard College},
	issn = {0017-8055, 1943-5045},
	shorttitle = {Undoing {Appropriateness}},
	url = {https://www.harvardeducationalreview.org/content/85/2/149},
	doi = {10.17763/0017-8055.85.2.149},
	language = {en},
	number = {2},
	urldate = {2026-04-22},
	journal = {Harvard Educational Review},
	publisher = {Harvard Educational Review},
	author = {Flores, Nelson and Rosa, Jonathan},
	month = jun,
	year = {2015},
	pages = {149--171},
}

@inproceedings{solyst_i_2023,
	address = {New York, NY, USA},
	series = {{CHI} '23},
	title = {“{I} would like to design”: {Black} girls analyzing and ideating fair and accountable {AI}},
	isbn = {978-1-4503-9421-5},
	shorttitle = {“{I} {Would} {Like} to {Design}”},
	url = {https://dl.acm.org/doi/10.1145/3544548.3581378},
	doi = {10.1145/3544548.3581378},
	urldate = {2026-04-21},
	booktitle = {Proceedings of the 2023 {CHI} {Conference} on {Human} {Factors} in {Computing} {Systems}},
	publisher = {Association for Computing Machinery},
	author = {Solyst, Jaemarie and Xie, Shixian and Yang, Ellia and Stewart, Angela E.B. and Eslami, Motahhare and Hammer, Jessica and Ogan, Amy},
	month = apr,
	year = {2023},
	pages = {1--14},
}

@misc{solyst_investigating_2025,
	title = {Investigating youth {AI} auditing},
	url = {http://arxiv.org/abs/2502.18576},
	doi = {10.48550/arXiv.2502.18576},
	urldate = {2026-04-22},
	publisher = {arXiv},
	author = {Solyst, Jaemarie and Peng, Cindy and Deng, Wesley Hanwen and Pratapa, Praneetha and Hammer, Jessica and Ogan, Amy and Hong, Jason and Eslami, Motahhare},
	month = feb,
	year = {2025},
	note = {arXiv:2502.18576 [cs]
version: 1},
}

@incollection{morales-navarro_building_2025,
	address = {New York, NY, USA},
	title = {Building {babyGPTs}: {Youth} engaging in data practices and ethical considerations through the construction of generative language models},
	isbn = {979-8-4007-1473-3},
	shorttitle = {Building {babyGPTs}},
	url = {https://dl.acm.org/doi/10.1145/3713043.3731525},
	urldate = {2026-04-21},
	booktitle = {Proceedings of the 24th {Interaction} {Design} and {Children}},
	publisher = {Association for Computing Machinery},
	author = {Morales-Navarro, Luis and Noh, Daniel J. and Kafai, Yasmin},
	month = jun,
	year = {2025},
	pages = {1021--1026},
}

@inproceedings{tanksley_ethics_2025,
	address = {New York, NY, USA},
	series = {{CHI} '25},
	title = {"{Ethics} is not neutral": {Understanding} ethical and responsible {AI} {Design} from the lenses of {Black} youth},
	isbn = {979-8-4007-1394-1},
	shorttitle = {"{Ethics} is not neutral"},
	url = {https://dl.acm.org/doi/10.1145/3706598.3713510},
	doi = {10.1145/3706598.3713510},
	urldate = {2026-04-21},
	booktitle = {Proceedings of the 2025 {CHI} {Conference} on {Human} {Factors} in {Computing} {Systems}},
	publisher = {Association for Computing Machinery},
	author = {Tanksley, Tiera and Smith, Angela D. R. and Sharma, Saloni and Huff, Earl W},
	month = apr,
	year = {2025},
	pages = {1--20},
}

@misc{charity_hudley_understanding_nodate,
	title = {Understanding {English} language variation in {U}.{S}. schools},
	url = {https://www.tcpress.com/products/understanding-english-language-variation-in-u-s-schools_9780807751480},
	language = {en},
	urldate = {2026-04-22},
	journal = {Teachers College Press},
	publisher = {Teachers College Press},
	author = {Charity Hudley, Anne H. and Mallinson, Christine},
}

@article{gholson_restoring_2019,
	title = {Restoring mathematics identities of {Black} learners: {A} curricular approach},
	volume = {58},
	issn = {0040-5841},
	shorttitle = {Restoring {Mathematics} {Identities} of {Black} {Learners}},
	url = {https://doi.org/10.1080/00405841.2019.1626620},
	doi = {10.1080/00405841.2019.1626620},
	number = {4},
	urldate = {2025-06-01},
	journal = {Theory Into Practice},
	publisher = {Routledge},
	author = {Gholson, Maisie L. and and Robinson, Darrius D.},
	month = oct,
	year = {2019},
	note = {\_eprint: https://doi.org/10.1080/00405841.2019.1626620},
	pages = {347--358},
}

@article{ortiz_black_2021,
	title = {Black {English} and mathematics education: {A} critical look at culturally sustaining pedagogy},
	volume = {123},
	url = {https://doi.org/10.1177/01614681211058978},
	doi = {10.1177/01614681211058978},
	number = {10},
	journal = {Teachers College Record},
	author = {Ortiz, Nickolaus Alexander and Ruwe, Dalitso},
	year = {2021},
	note = {\_eprint: https://doi.org/10.1177/01614681211058978},
	pages = {185--212},
}

@article{erath_designing_2021,
	title = {Designing and enacting instruction that enhances language for mathematics learning: a review of the state of development and research},
	volume = {53},
	issn = {1863-9704},
	shorttitle = {Designing and enacting instruction that enhances language for mathematics learning},
	url = {https://doi.org/10.1007/s11858-020-01213-2},
	doi = {10.1007/s11858-020-01213-2},
	language = {en},
	number = {2},
	urldate = {2026-04-22},
	journal = {ZDM – Mathematics Education},
	author = {Erath, Kirstin and Ingram, Jenni and Moschkovich, Judit and Prediger, Susanne},
	month = may,
	year = {2021},
	pages = {245--262},
}

@article{ortiz_lessons_2024,
	title = {Lessons in paradise: envisioning a {Black} liberatory mathematics education},
	volume = {116},
	issn = {1573-0816},
	shorttitle = {Lessons in paradise},
	url = {https://doi.org/10.1007/s10649-023-10263-8},
	doi = {10.1007/s10649-023-10263-8},
	language = {en},
	number = {3},
	urldate = {2026-04-22},
	journal = {Educational Studies in Mathematics},
	author = {Ortiz, Nickolaus Alexander},
	month = jul,
	year = {2024},
	pages = {539--550},
}

@article{morales_jr_underlife_2025,
	title = {The underlife of a mathematics classroom: {Latinx} bilinguals navigating the official and unofficial spaces},
	volume = {14},
	copyright = {http://creativecommons.org/licenses/by/4.0},
	issn = {2014-3621},
	shorttitle = {The {Underlife} of a {Mathematics} {Classroom}},
	url = {https://hipatiapress.com/hpjournals/index.php/redimat/article/view/16567},
	doi = {10.17583/redimat.16567},
	language = {en},
	number = {2},
	urldate = {2026-04-22},
	journal = {Journal of Research in Mathematics Education},
	author = {Morales, Jr., Hector and DiNapoli, Joseph},
	month = jun,
	year = {2025},
	pages = {115--138},
}

@article{poza_language_2018,
	title = {The language of ciencia: translanguaging and learning in a bilingual science classroom},
	volume = {21},
	issn = {1367-0050},
	shorttitle = {The language of ciencia},
	url = {https://doi.org/10.1080/13670050.2015.1125849},
	doi = {10.1080/13670050.2015.1125849},
	number = {1},
	urldate = {2026-04-22},
	journal = {International Journal of Bilingual Education and Bilingualism},
	publisher = {Routledge},
	author = {Poza, Luis E.},
	month = jan,
	year = {2018},
	note = {\_eprint: https://doi.org/10.1080/13670050.2015.1125849},
	pages = {1--19},
}

@article{martinez_puras_2014,
	title = {Puras groserías?: {Rethinking} the role of profanity and graphic humor in {Latin}@ students' bilingual wordplay},
	shorttitle = {Puras {Groserías}?},
	url = {https://www.academia.edu/43859911/Puras_Groser%C3%ADas_Rethinking_the_Role_of_Profanity_and_Graphic_Humor_in_Latin_at_Students_Bilingual_Wordplay},
	doi = {10.1111/AEQ.12074},
	urldate = {2026-04-22},
	journal = {Anthropology \& Education Quarterly},
	author = {Martínez, Ramón A. and Morales, P. Zitlali},
	month = jan,
	year = {2014},
}

@book{rosa_looking_2018,
	title = {Looking like a language, sounding like a race: {Raciolinguistic} ideologies and the learning of latinidad},
	shorttitle = {Looking like a {Language}, {Sounding} like a {Race}},
	url = {https://ccsre.stanford.edu/publications/looking-language-sounding-race-raciolinguistic-ideologies-and-learning-latinidad},
	language = {en},
	urldate = {2026-04-22},
	publisher = {Oxford University Press},
	author = {Rosa, Jonathan},
	year = {2018},
}

@incollection{planas_quality_2021,
	address = {Abingdon, Oxon ; New York, NY : Routledge, 2021.},
	edition = {1},
	title = {Quality dimensions for activation and participation in language-responsive mathematics classrooms},
	isbn = {978-0-429-26088-9},
	url = {https://www.taylorfrancis.com/books/9780429523007/chapters/10.4324/9780429260889-12},
	doi = {10.4324/9780429260889-12},
	language = {en},
	urldate = {2026-04-22},
	booktitle = {Classroom {Research} on {Mathematics} and {Language}},
	publisher = {Routledge},
	author = {Erath, Kirstin and Prediger, Susanne},
	editor = {Planas, Núria and Morgan, Candia and Schütte, Marcus},
	month = mar,
	year = {2021},
	pages = {167--183},
}

@article{webb_engaging_2014,
	title = {Engaging with others’ mathematical ideas: {Interrelationships} among student participation, teachers’ instructional practices, and learning {\textbar} {Request} {PDF}},
	shorttitle = {Engaging with others’ mathematical ideas},
	url = {https://www.researchgate.net/publication/259132689_Engaging_with_others'_mathematical_ideas_Interrelationships_among_student_participation_teachers'_instructional_practices_and_learning},
	doi = {10.1016/j.ijer.2013.02.001},
	language = {en},
	urldate = {2026-04-22},
	journal = {International Journal of Educational Research},
	author = {Webb, Noreen and Franke, Megan L. and Ing, Marsha and Wong, Jacqueline},
	month = dec,
	year = {2014},
}

\appendix
\section{Research Methods}

\subsection{Student Interviews}

\begin{table*}[h]
\centering
\caption{Overview of Participant Retrospection and Member Check Protocol}
\label{tab:membercheck}
\begin{tabular}{p{0.22\textwidth} p{0.68\textwidth}}
\toprule
\textbf{Phase} & \textbf{Questions / Prompts} \\
\midrule

Participant Retrospection Introduction 
& I want to talk about some of the moments when you were talking in your math class and how you felt and what you were thinking during those moments. Let’s read this example together.
 \\

\midrule

Retrospection Questions

& “Can you tell me more about this and what you were doing here?" \\
& “Could you tell me how you were feeling during this moment?” \\

\midrule

Member Check Introduction
& One of the things we want to be able to learn from looking at how students talk in math is how to help teachers notice the brilliant and interesting conversations you all are having.  As part of our research to understand the ways students talk in math class we are using Large Language Models (think of something like ChatGPT) to see how they pick up on the way you all talk to each other. So we can give the model a written version of your conversation and ask it to look for when a student asked a question, added on to what someone else said, or shared an idea about math. Here are some examples of what this model picked up from your conversation. Thinking about what you remember happening in this moment, what do you think about what the model noticed? \\

\midrule

Member Check Questions
& “If you agree, why do you agree?" \\
& "If you disagree, why do you disagree?” \\
& “Would you change anything or add on to what the model noticed?” \\
& “Was your intent captured correctly? Are there any things that the LLM is missing?” \\
& “Is there anything else you wish your teacher would notice when you are talking during math class?” \\

\midrule

Closing
& “Is there anything else you think I should know about math talk?”
\\

\bottomrule
\end{tabular}
\end{table*}

\end{document}